\documentclass[11pt]{article}
\usepackage{acl}
\usepackage[T1]{fontenc}
\usepackage{times}
\usepackage{latexsym}
\usepackage{amsmath,amssymb}
\usepackage{graphicx}
\usepackage{booktabs}
\usepackage{array}
\usepackage{tabularx}
\usepackage{adjustbox}
\usepackage{float}
\usepackage{placeins}
\usepackage{xurl}
\graphicspath{{./}}

\title{Are Diversity Metrics Measuring Diversity? A Capability-Controlled Audit of Majority-Vote Gain in LLM Ensembles}
\author{\mdseries Donghwan Kim \\
  Aidentyx Inc., San Jose, CA, USA \\
  \texttt{david.kim@aidentyx.com}}

\begin{document}
\maketitle
\begin{abstract}
Majority voting over LLMs is often expected to benefit from diversity, and diversity measures are often used to choose models to combine. We ask whether five such measures provide information about majority-vote gain beyond constituent capability. We study their associations with gain over the strongest member under explicit capability controls. We use three settings: MMLU-Pro ($30$ models, $453$ complete items), TruthfulQA ($22$ models, $400$ items; randomized option order), and MuSiQue ($20$ models, $578$ items; correctness-count diagnostic). We find three main results. First, oracle complementarity is common, but size-3 voting beats the strongest member in only $9.95\%$, $4.68\%$, and $0.61\%$ of ensembles. Second, strict diversity is nearly collinear with one minus mean accuracy ($\rho=+0.991/+0.988/+0.967$). An exact identity explains much of this relationship, while the observed strength depends on the model pool. Third, best+mean controlled associations depend on the metric and benchmark. Double-fault is negative on MMLU-Pro ($-0.398$) and TruthfulQA ($-0.515$), with model-subpool intervals excluding zero; the MuSiQue estimate is weaker, uncertain, and roster-sensitive ($-0.175$). Its size-3 link to gain is partly algebraic and becomes weaker under stronger capability adjustment. We contribute a capability-controlled measurement audit of diversity statistics for heterogeneous LLM voting.
\end{abstract}

\section{Introduction}

Majority voting is a standard aggregation rule in ensemble learning \citep{dietterich2000ensemble}. The main idea is simple: if base predictors fail on different inputs, a vote may correct individual errors and outperform one member \citep{kuncheva2003measures}. The rule is also common in LLM research, from self-consistency decoding \citep{wang2023selfconsistency} to multi-agent sampling-and-voting \citep{li2024more}. Diversity measures are often used to decide which models to combine.

Two facts make this question harder for modern LLMs. LLM errors are strongly correlated, and more accurate models can be \emph{more} correlated \citep{pmlr-v267-kim25e,pmlr-v267-goel25b}. Recent work balances competence and diversity through adaptive filtering \citep{cohen-etal-2026-dfpe}, quality-matched pools and co-failure ceilings \citep{chen2026combining}, or accuracy-adjusted dependence measures \citep{ali2026quantifying}. We ask a measurement question: \emph{for heterogeneous LLM pools, what do five diversity-related measures reveal about majority-vote gain over the strongest constituent, before and after explicit capability control?} We study broad multiple-choice knowledge and reasoning with MMLU-Pro, misconception-sensitive factuality with TruthfulQA, and document-grounded multi-hop free-form QA with MuSiQue. These complementary settings test whether shared measurement patterns recur and where conditional relationships depend on evaluation context. The rosters and protocols are not matched, so benchmark contrasts are descriptive rather than causal effects of task type.

We define realized majority-vote gain as the vote's accuracy minus the accuracy of the best individual member. We use it as the dependent variable and study five diversity-related measures under explicit capability controls. Beating the strongest available member is a strict criterion. Cost, latency, and model-selection uncertainty also matter in deployment \citep{chen2024frugalgpt}. However, if an ensemble cannot outperform its strongest constituent on accuracy, diversity alone does not justify aggregation on accuracy grounds. Our held-out analysis covers the case where the strongest member is not known in advance. We study unweighted correctness-count strict-majority voting over heterogeneous model subsets; optimized routing, learned selection, weighted voting, judge-based aggregation, and semantic answer synthesis are outside the study's estimand.

Oracle complementarity is present in $100\%$ of size-3 ensembles, but realized size-3 majority-vote gain is uncommon: $9.95\%$ on MMLU-Pro ($30$ models, $453$ complete items), $4.68\%$ on TruthfulQA ($22$ models, $400$ items; randomized option order), and $0.61\%$ on MuSiQue ($20$ models, $578$ items). Voting \emph{sometimes} beats the strongest member, but only in a minority of these ensembles.

Strict diversity is nearly collinear with $1-\overline{\mathrm{Acc}}$ ($\rho=+0.991/+0.988/+0.967$), a relationship partly fixed by the exact identity $\mathrm{Strict}=2(1-\overline{\mathrm{Acc}})-\mathrm{DF}$. After best+mean capability control, pairwise double-fault is negative ($-0.398/-0.515/-0.175$), but full-member-accuracy adjustment attenuates these associations to $-0.284/-0.204/-0.065$. These diagnostic and associational findings provide a capability-aware basis for interpreting diversity signals (Discussion).

Our contributions are:

\begin{enumerate}
\item \textbf{A capability-controlled measurement framework.} We audit associations between five diversity-related measures and realized majority-vote gain over the strongest constituent under nested capability controls, with model-level uncertainty, stronger-control sensitivity, and common-route robustness.
\item \textbf{Capability entanglement and algebraic clarification.} We quantify strong capability entanglement, especially for strict diversity, while using established identities among selected contingency-table measures ($\mathrm{strict}=\mathrm{disagreement}+\mathrm{double\text{-}fault}$; $\mathrm{strict}=2(1-\overline{\mathrm{Acc}})-\mathrm{double\text{-}fault}$) to separate algebraic constraints from empirical association magnitudes.
\item \textbf{Three-benchmark evidence of rare realized gain and control-dependent residuals.} Across MMLU-Pro, TruthfulQA, and MuSiQue, oracle complementarity rarely becomes realized majority-vote gain. Capability-adjusted associations depend on the metric, benchmark, and control specification. At size~3, double-fault must be interpreted through its exact relationship with gain rather than as an independent mechanism.
\end{enumerate}

Our claims concern measurement and association: capability controls estimate conditional relationships rather than causal effects, and their magnitudes depend on roster and item coverage. Section~4 describes the experimental setup; Appendix~\ref{app:historical-tqa} documents the material TruthfulQA protocol correction.

\section{Related Work}

\textbf{2.1 Classical ensemble diversity.} Classical ensemble learning established that diversity interacts with member accuracy, that maximizing diversity alone need not improve performance, and that pairwise diversity statistics can have weak or inconsistent relationships with ensemble accuracy \citep{dietterich2000ensemble,brown2005diversity,krogh1995neural,zhang2006ensemble,wood2023unified,kuncheva2003measures}. \citet{kuncheva2003measures} catalogued contingency-table measures and reported that their link to ensemble accuracy is often weak on real problems. Their catalogue defines disagreement and double-fault (their Eqs.~6--7), but not $(b{+}c{+}d)/N$ or Jaccard error overlap. Strict diversity is our name for the complement of joint correctness. Building on this accuracy--diversity literature, we provide an LLM-specific, metric-by-metric study of \emph{gain over the strongest constituent}, with explicit capability adjustment and model-level uncertainty on heterogeneous modern LLM pools.

\textbf{2.2 Voting and self-consistency in LLMs.} Voting-style aggregation is widely used with LLMs: self-consistency decoding samples multiple reasoning paths and selects the modal answer \citep{wang2023selfconsistency}, extended to free-form generation by universal self-consistency \citep{chen2024universal}, and multi-agent studies report gains from sampling-and-voting as the number of instantiated agents grows \citep{li2024more,pmlr-v235-du24e}. These results motivate majority voting as an object of study, but they largely concern single-model self-sampling gains over an agent's own baseline, not when voting over a \emph{heterogeneous} pool beats the strongest available member. Condorcet-style guarantees require stronger conditions than is sometimes stated: independence of classifier outputs alone is not enough without suitable independence of their errors \citep{vardeman2013majority}. In financial-sentiment experiments, heterogeneous NLP/LLM voting produced only marginal improvement, which the authors attribute to insufficient independence \citep{lefort2024examining}. This is task-specific preprint evidence, not a general theorem about LLMs. Our question is the measurement version: when majority-vote gain over the best member is the target for \emph{heterogeneous} LLM subsets, do diversity-related measures carry information beyond member capability?

\textbf{2.3 Correlated errors and model similarity.} \citet{pmlr-v267-kim25e} show LLM errors are correlated and that stronger models are more correlated; \citet{pmlr-v267-goel25b} find stronger models make more similar mistakes. These works document the \emph{empirical pattern} behind our capability confound, in which accurate models agree and fail together. However, they study error correlation itself, not majority-vote gain, and do not measure how strongly standard diversity measures re-express capability. Our analysis studies this consequence for diversity measurement.

\textbf{2.4 LLM ensemble selection, quality control, and aggregation.} A growing literature \emph{designs} aggregation and selection rules for LLMs \citep{chen2025harnessing}, including ranking and fusion \citep{jiang-etal-2023-llm,si-etal-2023-getting}, layered proposer--aggregators \citep{wang2025mixture,li2026rethinking}, focal-diversity pruning \citep{tekin-etal-2024-llm}, competence filtering \citep{cohen-etal-2026-dfpe}, higher-order aggregation \citep{ai2026beyond}, error-distribution shaping \citep{baek2026dfuser}, and complementarity- or information-theoretic selection \citep{zhang2026complementary,turkmen2026highest,teng2026whichmodels}. Concurrent work on all-member co-failure ceilings, model-pool selection for multi-agent systems, and accuracy-adjusted dependence converges on two premises of our audit: combining models rarely beats the strongest member \citep{chen2026combining,marjanovic2026momodels}, and raw dependence can mislead when capability differs \citep{chen2026combining,ali2026quantifying}. We complement these methods with a descriptive, metric-specific audit of five measures as conditional correlates of realized unweighted majority-vote gain.

\section{Preliminaries}

\subsection{Contingency-table measures, exact identities, and focal diversity}

For a pair of models on $N$ items, the standard $2\times2$ contingency table \citep{kuncheva2003measures} assigns each item to one cell by whether each model is correct:

\begin{center}
\small
\begin{tabular}{lll}
\toprule
 & \emph{j} correct & \emph{j} wrong \\
\midrule
\textbf{i correct} & \emph{a} & \emph{b} \\
\textbf{i wrong} & \emph{c} & \emph{d} \\
\bottomrule
\end{tabular}
\end{center}

We select five measures to span joint correctness, split errors, shared errors, normalized error overlap, and a recent LLM-specific subset criterion. The audit is not intended as an exhaustive catalogue of classical diversity indices.

\begin{figure*}[t]
\centering
\includegraphics[width=\textwidth]{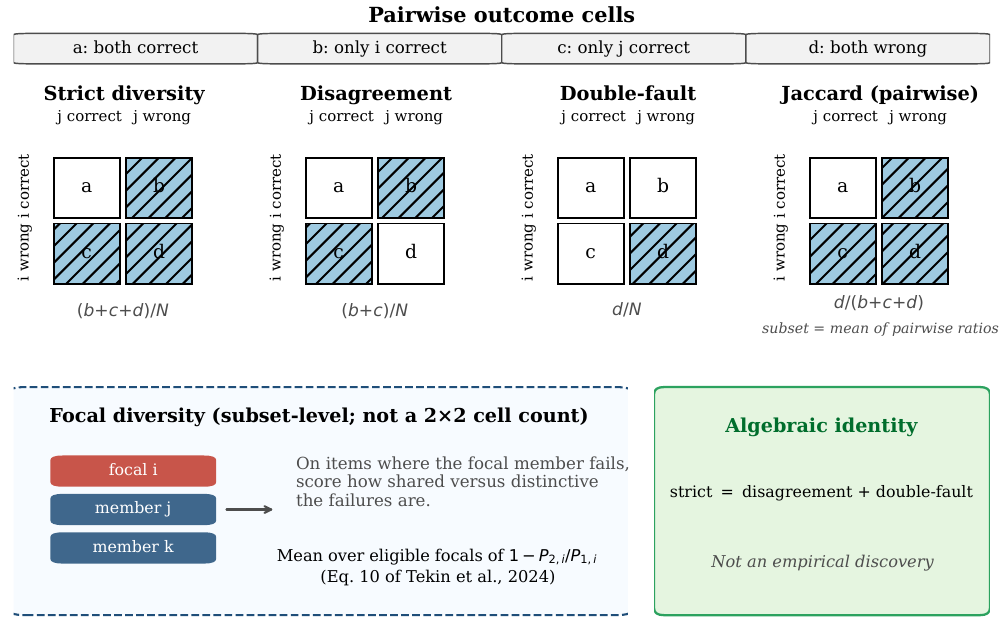}
\caption{Intuitive overview of the five audited measures. Top: four pairwise contingency-table summaries over cells $b$ (only $i$ correct), $c$ (only $j$ correct), and $d$ (both wrong), with subset Jaccard defined as the mean of pairwise ratios $d/(b{+}c{+}d)$. Bottom-left: focal diversity, a separate subset-level failure-structure score for a focal member \citep[Eq.~10]{tekin-etal-2024-llm}, not a fifth cell count of the same $2\times2$ table. Bottom-right: the exact algebraic identity $\mathrm{strict}=\mathrm{disagreement}+\mathrm{double\text{-}fault}$. Capability entanglement and associations with majority-vote gain are empirical questions addressed in Section~5.}
\label{fig:metric-overview}
\end{figure*}

Figure~\ref{fig:metric-overview} summarizes the roles of the five audited statistics before the formal definitions. Four are classical pairwise contingency-table summaries of whether models $i$ and $j$ are correct or wrong on each of $N$ items. Cells $b$ and $c$ are split outcomes (exactly one model wrong), and cell $d$ is shared failure. Strict diversity $(b{+}c{+}d)/N$ asks how often the pair is \emph{not jointly correct}; disagreement $(b{+}c)/N$ counts split outcomes; double-fault $d/N$ counts shared failures; and pairwise Jaccard error similarity $d/(b{+}c{+}d)$ divides shared failure by the union of errors. For a model subset, we average the pairwise Jaccard \emph{ratios}, not a pooled ratio of summed counts. These four measures are not interchangeable. The exact identity $\mathrm{strict}=\mathrm{disagreement}+\mathrm{double\text{-}fault}$ is algebraic. Whether a statistic mainly re-expresses member capability, and whether its residual association with majority-vote gain remains after capability controls, are empirical questions addressed in Section~5. Focal diversity is shown separately. It is a subset-level failure-structure score for a focal member relative to the group \citep[Eq.~10]{tekin-etal-2024-llm}, not another cell count of the same pairwise table.

In particular, the Q-statistic, the correlation coefficient, and kappa are functions of the same $2\times2$ cell counts, and \citet{kuncheva2003measures} report that these pairwise measures are strongly intercorrelated; we therefore select representatives spanning distinct structural roles rather than near-duplicates.

We audit four contingency-table measures, extended to subsets of size $k$ by averaging over member pairs:

\begin{itemize}
\item \textbf{Strict diversity} (our term; not a named measure in Kuncheva and Whitaker's catalogue): $\mathrm{SD} = (b+c+d)/N$, the complement of joint correctness.
\item \textbf{Disagreement:} $(b+c)/N$
\item \textbf{Double-fault (co-failure):} $d/N$
\item \textbf{Mean pairwise Jaccard error-set similarity:} $d/(b+c+d)$, the standard Jaccard index on model error sets, averaged over pairs (0 when both error sets are empty). The mean of pairwise ratios is close to, but not identical with, the pooled ratio of sums (max divergence 0.089); we retain the pairwise-mean definition throughout.
\end{itemize}

Two exact identities frame the analysis. First,
\begin{equation*}
\begin{aligned}
\text{strict} &= \text{disagreement} + \text{double-fault},\\
\max |\epsilon| &\leq 3\times 10^{-16},
\end{aligned}
\end{equation*}
where $\epsilon$ is the numerical verification error. Thus, strict, disagreement, and double-fault are linearly dependent, both pairwise and after pairwise averaging. At the pair level, Jaccard is the ratio $\mathrm{double\text{-}fault}/\mathrm{strict}$. Because the subset statistic averages pairwise Jaccard ratios, however, it is not algebraically determined by the subset-level averaged double-fault and strict measures (the maximum observed divergence from their ratio is 0.089). Second, with $\overline{\mathrm{Acc}}=(2a+b+c)/(2N)$ the pair mean accuracy,
\begin{equation*}
1-\overline{\mathrm{Acc}} = \mathrm{DoubleFault} + \tfrac{1}{2}\mathrm{Disagreement},
\end{equation*}
an exact identity preserved under pairwise averaging to the subset level. Combining the two identities gives
\begin{equation*}
\mathrm{Strict}=2(1-\overline{\mathrm{Acc}})-\mathrm{DoubleFault},
\end{equation*}
for any subset size under these uniform pairwise-mean definitions. In raw (unranked) measure space, any linear residualization that includes mean member accuracy therefore forces
\begin{equation*}
\mathrm{DoubleFault}_{\mathrm{res}} = -\tfrac{1}{2}\mathrm{Disagreement}_{\mathrm{res}},
\end{equation*}
exactly (Pearson $r=-1$, slope $-1/2$). This identity need not be preserved after the marginal rank transformations used for partial Spearman correlations, so the controlled magnitudes in Appendix Table~\ref{tab:five-metric-associations} need not be symmetric. Raw-space one-dimensionality after mean control is thus \emph{algebraic}; whether the corresponding rank-space association with majority-vote gain is robust is \emph{empirical} (Sections~5.4--5.5).

Separately, we audit focal diversity \citep[Eq.~10]{tekin-etal-2024-llm}, a subset-level measure that is \emph{not} a member of this algebraic family. For a subset of $k$ models, let $F_q$ be the number of members that fail on item $q$. For each focal member $i$ that fails at least once, we compute
\begin{equation*}
\begin{aligned}P_{1,i}&=\mathbb{E}_{q:i\text{ fails}}\!\left[\frac{F_q}{k}\right],\\ P_{2,i}&=\mathbb{E}_{q:i\text{ fails}}\!\left[\frac{F_q(F_q-1)}{k(k-1)}\right],\\ \rho_i&=1-\frac{P_{2,i}}{P_{1,i}},\end{aligned}
\end{equation*}
and define subset focal diversity as the mean of $\rho_i$ over eligible focal members (undefined focal terms are skipped). Focal is not one of the three linearly coupled contingency-table summaries above.

\subsection{Majority vote, gain, and an oracle upper bound}

For subset $S$ of size $k$, majority vote counts an item correct iff the number of correct members strictly exceeds $k/2$ (a correct-count rule; exact even-size splits count as incorrect). The same correctness-count majority rule is the primary voting estimand on MuSiQue; there the count is an oracle-style diagnostic over free-form correctness outcomes, not a deployable vote over answer strings. Appendix~C reports a MuSiQue sensitivity using answer-level plurality (most-voted normalized answer wins). With $\mathrm{Acc}_{\mathrm{MV}}(S)$ and $\mathrm{Acc}_{\mathrm{best}}(S)$ the best member's accuracy on the same item set:

\begin{itemize}
\item \textbf{Gain:} $g(S) = \mathrm{Acc}_{\mathrm{MV}}(S) - \mathrm{Acc}_{\mathrm{best}}(S)$.
\item \textbf{Oracle upper bound:} the gain of a hypothetical aggregator that is correct whenever \emph{any} member is. This is not a real aggregator and is used only as an upper bound on potential complementarity.
\end{itemize}

At size~3, let $B=\mathrm{Acc}_{\mathrm{best}}$, let $D$ be mean pairwise double-fault, and let $F$ be the fraction of items on which all three members fail. Exact counting gives
\begin{equation}
G = 1-B-3D+2F.
\label{eq:size3-gain}
\end{equation}
This established size-3 bookkeeping identity connects classical majority-vote decompositions \citep{kuncheva2003limits,brown2010goodbad} to recent co-failure and rescue/damage analyses \citep{chen2026combining,ali2026quantifying}. It means that a negative conditional double-fault association is partly structurally expected, while its rank-based magnitude and the offset from $F$ remain empirical.

Under the correct-count rule, size-2 majority vote can never exceed the best member (MV correct requires both correct), so size 2 is excluded from gain-rate headlines. Unless stated otherwise, ``best'' is selected in-sample; Section~5.1 quantifies held-out selection. All associations are between a measure and $g(S)$ across subsets.

\section{Experimental Setup}

\textbf{Benchmarks.} We evaluate three frozen samples that vary in task demand and answer format. MMLU-Pro \citep{wang2024mmlupro} covers broad multiple-choice knowledge and reasoning (500 items sampled with seed 42). TruthfulQA \citep{lin-etal-2022-truthfulqa} probes factuality and common misconceptions (500 multiple-choice items sampled with seed 42). MuSiQue \citep{trivedi-etal-2022-musique} requires document-grounded multi-hop reasoning with free-form answers (600 hop-stratified items: 300/180/120 two-/three-/four-hop). This selection tests whether the same measurement relationships recur across meaningfully different formats and model pools, while allowing context dependence. The settings are not a difficulty ordering or matched experimental conditions, and observed differences cannot identify task-format effects.

\textbf{Model roster design.} Inference uses OpenRouter routes, with default provider routing rather than pinned providers. The benchmark-specific primary rosters contain 30 MMLU-Pro routes meeting parse rate $\geq0.90$ and the pragmatic full-sample accuracy band $[0.40,0.92]$; 22 available original TruthfulQA routes meeting the randomized parse-rate criterion; and a fixed 20-route MuSiQue roster. Different route availability and construction criteria explain the 30/22/20 counts. The exact 19-route intersection provides a partial roster-composition check but does not align items, prompts, dates, answer format, or scoring (Appendix~\ref{app:shared-roster}). Tables~A1--A3 report route identifiers, benchmark-specific criteria, unavailable routes, and roster sensitivities.

\textbf{Sampling and prompts.} MMLU-Pro and TruthfulQA use task-specific zero-shot chain-of-thought prompts with no demonstrations; each requests a final option label. TruthfulQA options are independently permuted with seed 2026, with gold positions balanced within each choice-count group. MuSiQue prompts include the question and all provided supporting and distractor paragraphs and request a marked final answer. Exact templates and generation settings are in Appendix~A.2. Collection dates, model versions, and serving conditions were not uniform across all benchmarks or routes.

\textbf{Parsing and correctness.} Multiple-choice responses are mapped to option labels only when a valid final label is identified. The TruthfulQA parser applies a uniform, gold-blind standalone-label criterion: it retains valid standalone labels, re-parses completed cue-plus-word responses with stricter boundaries, leaves truncated or final-answer-invalid responses missing, and never creates a parse for a response that was originally unparsed. Truncated responses without a valid final label follow a failure-based output-token ladder (1,024$\rightarrow$4,096$\rightarrow$8,192); this produces nonuniform budgets across items and routes. MuSiQue correctness is maximum official normalized exact match over the answer and aliases. Missing predictions are not treated as incorrect.

\textbf{Ensembles and voting.} We enumerate every size-2--4 subset within each roster (MMLU-Pro: $435{+}4{,}060{+}27{,}405$; TruthfulQA: $231{+}1{,}540{+}7{,}315$; MuSiQue: $190{+}1{,}140{+}4{,}845$). The common estimand is correctness-count strict-majority gain over the strongest constituent on the same items; for MuSiQue, this is the diagnostic defined in Section~3.2.

\textbf{Capability controls and uncertainty.} The canonical estimate is a partial Spearman association obtained by residualizing ranked gain and the ranked diversity measure on ranked best-member and mean-member accuracy. Stronger specifications include the full sorted member-accuracy vector. Because enumerated subsets overlap, we do not use subset-level significance tests. We report direction and effect size, unique-model subpool robustness intervals ($B=1{,}000$; seed 0), and leave-one-model-out stability. The resampling procedure perturbs the model roster and excludes item-sampling uncertainty (Appendix~B).

\textbf{Complete-case analysis.} Primary analyses use the intersection of items with a valid prediction for every model in each final roster (Table~\ref{tab:primary-coverage}). The resulting matrices are MMLU-Pro $30{\times}453$, TruthfulQA $22{\times}400$, and MuSiQue $20{\times}578$. Missingness is concentrated by model and item rather than missing completely at random; Section~5.6 and the Limitations therefore treat complete-case selection as a scope limitation rather than a random subsample assumption.

\begin{table}[ht]
\centering
\small
\caption{Primary evaluation coverage.}
\label{tab:primary-coverage}
\begin{adjustbox}{max width=\columnwidth}
\begin{tabular}{lrrr}
\toprule
Benchmark & Models & Frozen & Complete \\
\midrule
MMLU-Pro & 30 & 500 & 453 \\
TruthfulQA & 22 & 500 & 400 \\
MuSiQue & 20 & 600 & 578 \\
\bottomrule
\end{tabular}
\end{adjustbox}
\end{table}

\textbf{Artifacts.} Primary results are computed from the three complete-case correctness matrices with common estimators. We plan to release the binary matrices, item IDs, rosters, and analysis scripts; raw prediction caches remain conditional on redistribution review.

\section{Results}

\emph{Estimator convention.} Unless noted, Spearman associations use the full size-3 subset population on each primary complete-case matrix. Capability-controlled columns are partial Spearman correlations: Pearson correlations of rank residuals after ordinary least-squares residualization on the stated ranked covariates (canonical control: best-member and mean member accuracy). Model-level uncertainty uses unique-model subpool percentile intervals ($B{=}1{,}000$; seed 0; Appendix~B), not ensemble-level bootstrap intervals. The full numeric five-metric matrix and stronger-control sensitivity appear in Appendix~C.

\subsection{Majority voting rarely beats the best member}

\textbf{Finding 1.} Oracle complementarity is present in every size-3 ensemble on all three primary benchmarks: for every triplet, at least one member is correct on some item where the strongest member is wrong (oracle-positive rate $100\%$ on MMLU-Pro, TruthfulQA, and MuSiQue). However, strict-majority gain over the strongest constituent is uncommon. Under the standard correctness-count rule with in-sample best selection, size-3 majority voting beats the strongest member in $404/4060$ ensembles on MMLU-Pro ($9.95\%$), $72/1540$ on TruthfulQA ($4.68\%$), and $7/1140$ on MuSiQue ($0.61\%$; Table~\ref{tab:three-benchmark-summary}). Mean realized gain is negative in each setting ($-0.037$, $-0.027$, and $-0.071$). Mean harm (the best member is correct while both others are wrong) is $2.72/3.30/3.93$ times mean rescue (the best is wrong while both others are correct). This helps explain why oracle opportunities often fail to become 2-of-3 improvements. Voting \emph{sometimes} improves on the best member, but only in a minority of these pools. Size~2 yields a structural zero under the same rule and is not evidence about diversity. Size~4 yields an empirical zero on these matrices under strict majority with ties counted as incorrect. We report size~3 as the main rate and do not pool sizes~2--4 into a single headline percentage.

\begin{table}[t]
\centering
\small
\begin{tabular}{@{}lrrr@{}}
\toprule
Benchmark & Gain $>0$ & Strict $\rho$ & DF $\rho$ \\
\midrule
MMLU-Pro & 9.95\% & 0.991 & $-0.398$ \\
TruthfulQA & 4.68\% & 0.988 & $-0.515$ \\
MuSiQue & 0.61\% & 0.967 & $-0.175$ \\
\bottomrule
\end{tabular}
\caption{Size-3 results on the primary matrices. Gain $>0$ is the percentage of triplets whose correctness-count majority beats the strongest member. Strict $\rho$ correlates strict diversity with one minus mean accuracy; DF $\rho$ is the best+mean partial Spearman association. Model-subpool intervals are reported in Section~5.4.}
\label{tab:three-benchmark-summary}
\end{table}

When the strongest member is instead selected on a random half of items and gain is evaluated on the held-out half (20 seeds; no evaluation-item leakage into selection), size-3 positive-gain rates rise but remain a minority outcome (Table~\ref{tab:heldout-size3}). The rise is expected: the member selected as strongest on the development half need not be strongest on the held-out half, so the held-out comparison baseline is often weaker than the in-sample maximum. Primary in-sample rates are unchanged by this check.

\begin{table}[t]
\centering
\small
\begin{tabular}{@{}lcc@{}}
\toprule
Benchmark & In-sample & Held-out \\
\midrule
MMLU-Pro & 9.95\% & $16.15\%\pm 2.79$ \\
TruthfulQA & 4.68\% & $12.81\%\pm 2.44$ \\
MuSiQue & 0.61\% & $3.42\%\pm 1.03$ \\
\bottomrule
\end{tabular}
\caption{Size-3 gain$>0$ rates under in-sample versus held-out best-member selection (seeds $0$--$19$). Held-out values are mean$\pm$SD; MuSiQue uses the diagnostic defined in Section~3.2.}
\label{tab:heldout-size3}
\end{table}

\begin{figure*}[t]
\centering
\includegraphics[width=\textwidth]{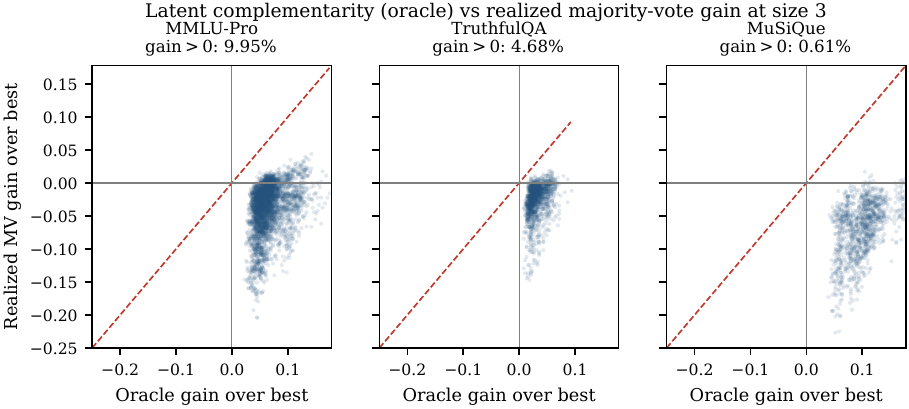}
\caption{Oracle opportunity versus realized majority-vote gain at size~3 on the verified primary matrices (MMLU-Pro $30{\times}453$, TruthfulQA $22{\times}400$, MuSiQue $20{\times}578$). Most ensembles lie below $y{=}0$ despite strictly positive oracle gain (shares above zero in panel titles). Dashed line: $y{=}x$. Overlapping ensembles are dependent.}
\label{fig:oracle-vs-realized}
\end{figure*}

The conclusion is the same in all three settings. Oracle opportunities are common, but simple correctness-count voting beats the strongest member in only a small minority of size-3 ensembles under either in-sample or held-out best selection (Figure~\ref{fig:oracle-vs-realized}).

\subsection{Raw measure--gain correlations are misleading}

Raw Spearman associations between each audited measure and size-3 gain already show why unadjusted rankings can mislead (Figure~\ref{fig:raw-controlled-heatmap}, left; numeric values in Appendix Table~\ref{tab:five-metric-associations}).

\begin{figure*}[t]
\centering
\includegraphics[width=\textwidth]{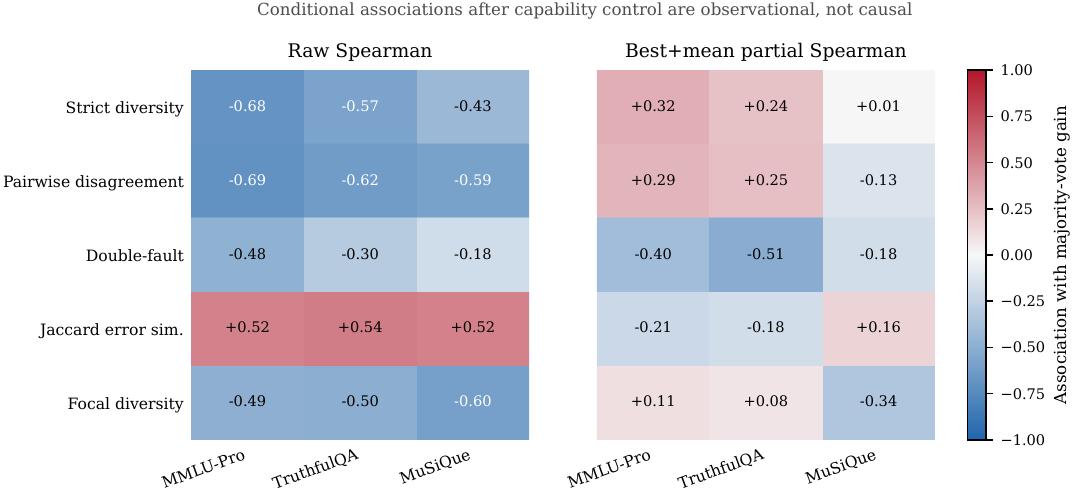}
\caption{Size-3 associations between diversity measures and correctness-count gain. Left: raw Spearman. Right: partial Spearman after residualizing ranks on best and mean member accuracy. Values from the verified three-benchmark matrices; overlapping ensembles are dependent. Conditional associations are observational, not causal.}
\label{fig:raw-controlled-heatmap}
\end{figure*}

Across all three benchmarks, the raw associations do not yield a uniform classical diversity-benefit pattern: strict diversity and pairwise disagreement are negatively associated with gain (MMLU-Pro: $-0.675$ and $-0.690$; TruthfulQA: $-0.571$ and $-0.621$; MuSiQue: $-0.425$ and $-0.585$), while double-fault is also negative ($-0.482$, $-0.299$, $-0.182$). Mean pairwise Jaccard error-set similarity is positively associated with gain in the raw data ($+0.517$, $+0.544$, $+0.518$), while focal diversity is negative ($-0.486$, $-0.499$, $-0.598$). Taken at face value, overlap-based proxies can even appear to carry the \emph{wrong sign}. The next subsections show that this reading is strongly shaped by capability entanglement, and that the five measures are not interchangeable once capability is controlled.

\subsection{Capability explains most measure variation}

\textbf{Finding 2.} Strict diversity is empirically near-collinear with capability in these pools:
\begin{equation*}
\begin{gathered}
\text{strict diversity} \approx 1-\text{mean member accuracy},\\
\begin{aligned}
&\rho_{\text{MMLU-Pro}}=+0.991,\ 
\rho_{\text{TruthfulQA}}=+0.988,\\
&\rho_{\text{MuSiQue}}=+0.967.
\end{aligned}
\end{gathered}
\end{equation*}
The exact identity $\mathrm{Strict}=2(1-\overline{\mathrm{Acc}})-\mathrm{DoubleFault}$ makes a substantial part of this capability relationship structural. The observed Spearman coefficients are not mathematically fixed: their strength depends on how mean accuracy and double-fault vary jointly in each heterogeneous model pool. Empirically, they manifest as the near-collinearities above, consistent with strongly correlated errors among modern LLMs \citep{pmlr-v267-kim25e}. A high raw ``diversity'' score can therefore largely reflect weaker constituents rather than independently useful complementarity.

Capability entanglement is strong for strict diversity but not uniform across the five measures. Spearman correlations with mean member accuracy are near $-1$ for strict diversity ($-0.991$, $-0.988$, $-0.967$) and remain strong for double-fault ($-0.881$, $-0.812$, $-0.934$). Pairwise disagreement is strongly tied to mean accuracy on the multiple-choice benchmarks ($-0.932$, $-0.916$) and more moderately on MuSiQue ($-0.546$). Jaccard error similarity and focal diversity are only moderately associated with mean accuracy on MMLU-Pro ($+0.455$ and $-0.455$) and TruthfulQA ($+0.392$ and $-0.330$); on MuSiQue, Jaccard is weak ($+0.021$) while focal is modest ($-0.173$). We therefore do not treat the five measures as interchangeable capability proxies, even though the three contingency-table statistics are algebraically coupled.

\textbf{Finding 3.} Under the canonical best+mean control, associations are metric- and benchmark-dependent (Figure~\ref{fig:raw-controlled-heatmap}, right; Appendix Table~\ref{tab:five-metric-associations}). Controlled strict diversity is $+0.324/+0.239/+0.006$; disagreement is $+0.290/+0.246/-0.130$; Jaccard is $-0.209/-0.179/+0.163$; focal diversity is $+0.106/+0.085/-0.335$; and double-fault is $-0.398/-0.515/-0.175$. Model-subpool intervals (Appendix~B) exclude zero for strict diversity, disagreement, and double-fault on MMLU-Pro, for double-fault on TruthfulQA, and for focal diversity on MuSiQue; TruthfulQA disagreement is borderline (lower endpoint $-0.0001$), and the remaining intervals include zero. These roster-perturbation intervals are not confidence intervals over independent subsets. Section~5.4 interprets the structural constraint on double-fault, and Section~5.5 tests stronger capability control.

\textbf{Cross-benchmark contrast.} MMLU-Pro has the largest observed positive-gain share ($9.95\%$). TruthfulQA has less frequent gain ($4.68\%$) but the strongest negative best+mean double-fault estimate ($-0.515$). MuSiQue has almost no positive correctness-count majority gain ($0.61\%$); its double-fault estimate is weaker and uncertain ($-0.175$, interval includes zero), while focal diversity provides a contrasting controlled association ($-0.335$, model-subpool interval excluding zero). Strict diversity remains tightly capability-entangled in every setting ($0.991/0.988/0.967$). These contrasts show both a shared measurement problem and setting dependence; because the rosters, items, prompts, answer formats, scoring, and serving conditions differ, they do not establish causal task effects.

\subsection{Interpreting residual pairwise co-failure}

The contingency-table statistics do not provide several independent linear signals. From the exact identity in Section~3.1, $1-\overline{\mathrm{Acc}}=\mathrm{DoubleFault}+\tfrac12\mathrm{Disagreement}$, any linear residualization of the raw (unranked) measures that includes mean member accuracy \emph{forces}
\begin{equation*}
\mathrm{DoubleFault}_{\mathrm{res}} = -\tfrac{1}{2}\mathrm{Disagreement}_{\mathrm{res}},
\end{equation*}
exactly, hence Pearson $r=-1.000$ with slope $-1/2$ (max numerical error $\leq3\times10^{-16}$; Appendix~B.1). This raw-space one-dimensionality follows algebraically. Appendix Table~\ref{tab:five-metric-associations} instead uses marginal ranks followed by linear residualization. Rank transformation does not preserve the raw linear identity, so the partial-Spearman magnitudes for disagreement and double-fault need not be symmetric ($+0.290$ vs $-0.398$ on MMLU-Pro). A raw-space joint regression of gain on \{disagreement, double-fault, mean\} is rank-deficient by construction.

\begin{figure*}[t]
\centering
\includegraphics[width=\textwidth]{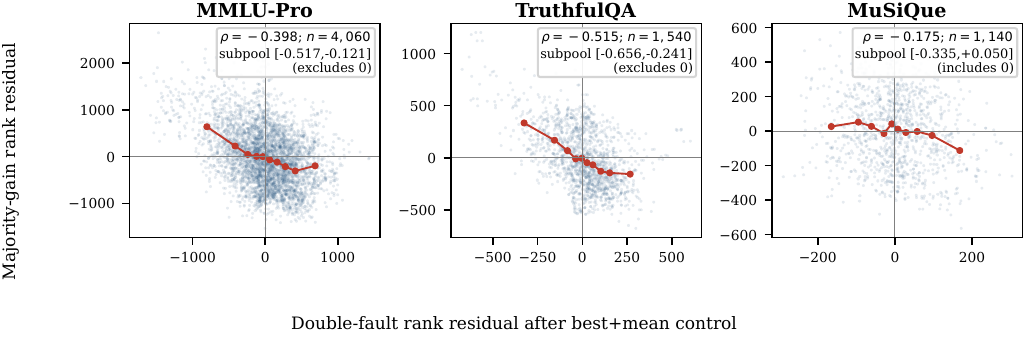}
\caption{Residual pairwise co-failure associations at size~3: rank residuals of gain versus double-fault after best+mean control, with decile-binned means. Each panel reports the controlled partial Spearman estimate, number of distinct triplets, and unique-model subpool interval. Axes retain benchmark-specific residual ranges; they should not be compared as common scales. MuSiQue's interval includes zero and its estimate is roster-sensitive. Points are overlapping ensembles and are not independent.}
\label{fig:partial-residual}
\end{figure*}

Equation~\ref{eq:size3-gain} further constrains the outcome: conditional on $B$, double-fault enters raw gain with coefficient $-3$, offset by $+2F$. A negative double-fault--gain association is therefore partly expected from vote mechanics, not evidence of an independent causal mechanism. The empirical quantities are the rank-based conditional magnitude and how the all-member-failure term covaries with double-fault in each pool. Under the best+mean rank control, the estimates are $-0.398$ on MMLU-Pro with model-resampling interval $[-0.517,-0.121]$, $-0.515$ on TruthfulQA with $[-0.656,-0.241]$, and $-0.175$ on MuSiQue with $[-0.335,+0.050]$ (Figure~\ref{fig:partial-residual}).

Leave-one-model-out ranges stay negative on MMLU-Pro and TruthfulQA ($[-0.422,-0.340]$ and $[-0.546,-0.424]$). MuSiQue's interval includes zero, so we do \emph{not} treat the association as equally robust across all three benchmarks. These roster-perturbation intervals do not include item-sampling uncertainty.

\subsection{Robustness and limits}

The best+mean estimates change under stronger capability adjustment. Controlling for the full sorted vector of all three member accuracies reduces double-fault to $-0.284$ on MMLU-Pro, $-0.204$ on TruthfulQA, and $-0.065$ on MuSiQue. The corresponding model-subpool intervals are $[-0.461,-0.049]$, $[-0.372,-0.053]$, and $[-0.310,+0.242]$ (Appendix Table~\ref{tab:full-vector-control}). The point estimates remain negative, but their magnitude depends on the control. The reduction is modest on MMLU-Pro, large on TruthfulQA, and the MuSiQue interval includes zero. The TruthfulQA full-vector estimate is also sensitive to parsing and missing-data choices. Its interval includes zero under an independent answer parser ($-0.157$) or when every truncated response is treated as missing ($-0.123$), so we do not treat it as stable (Appendix~\ref{app:historical-tqa}). Leave-one-model-out ranges under best+mean remain negative on all three matrices (MuSiQue: $[-0.212,-0.103]$). These checks treat overlapping ensembles as dependent and do not show causal or future selection utility.

By contrast, the canonical TruthfulQA best+mean estimate stays between $-0.49$ and $-0.52$ across parser, truncation, and single-route checks, whereas exact gain rates and the stronger full-vector estimate are more protocol-sensitive. Appendix~\ref{app:historical-tqa} reports the complete sensitivity table and output-budget rerun diagnostic.

An exploratory 20-split held-out analysis (Appendix~\ref{app:predictive-validity}) found that the algebraically coupled contingency-table measures provide one shared linear signal after mean-accuracy control and modestly improve out-of-item gain prediction beyond best and mean member accuracy ($\Delta R^2 \leq 0.061$), but under this Ridge specification none of the five metrics retains material incremental predictive value beyond the full sorted member-accuracy vector.

\subsection{Is the parsed slice driving the results?}

Complete-case filtering may remove items on which diversity would matter most. Headline associations use the complete-case intersections in Table~\ref{tab:primary-coverage}. Missing cells are never scored as incorrect (Section~4). Coverage is incomplete and concentrated rather than missing completely at random.

Fixed-roster diagnostics show that retained complete-case items are easier than excluded items (retained-minus-excluded accuracy $+0.186$ on MMLU-Pro, $+0.193$ on TruthfulQA over the 13 routes with full coverage, and $+0.300$ on a 12-model MuSiQue roster). On TruthfulQA, missingness is concentrated in a few routes, and a per-ensemble available-item analysis is reported in Appendix~\ref{app:historical-tqa}. For MMLU-Pro, an analogous size-3 analysis uses each triplet's observed intersection among the frozen 500 items. It gives a positive-gain rate of $7.78\%$ (versus $9.95\%$ on the roster-wide 453-item complete case) and controlled double-fault $-0.413$ (versus $-0.398$), with denominators from 460 to 500 (median 498). Because this is a different estimand, we use it only to assess sensitivity. It preserves the qualitative conclusions of Findings~1 and~3. We therefore treat missingness as a scope limitation, not as a missing-at-random assumption, and keep primary claims tied to the complete-case matrices. MuSiQue retains $299/175/104$ two-/three-/four-hop items; hop-wise diversity--gain associations were not validated.

\section{Discussion}

Capability strongly shapes the raw correlations suggesting that ``diversity hurts and overlap helps.'' Several audited measures are related to member accuracy, and strict diversity is nearly collinear with $1-\overline{\mathrm{Acc}}$ across the three primary pools. This measurement finding is consistent with competence filtering \citep{cohen-etal-2026-dfpe}, quality matching and all-member co-failure ceilings \citep{chen2026combining}, and accuracy-adjusted dependence \citep{ali2026quantifying}. The five statistics are not interchangeable. The contingency-table measures are algebraically coupled, while Jaccard and focal diversity have weaker or less consistent associations with capability. Capability-controlled associations ask whether a measure retains \emph{conditional} information about majority-vote gain given observed best and mean accuracy. They do not isolate a causal effect of diversity.

Intervention work such as D-FUSEr shows that changing error distributions can improve majority vote \citep{baek2026dfuser}. Our claim is narrower: observational diversity statistics over heterogeneous pools are strongly related to capability, especially for the linearly coupled contingency-table trio. Future work may need error-shaping interventions or capability-controlled measurement instead of relying only on raw diversity scores.

The negative direction of a residual co-failure association is partly expected from Equation~\ref{eq:size3-gain}. The sign alone is not the empirical contribution. The main finding is that its magnitude varies with the benchmark and capability specification, while Jaccard and focal behave differently from the linearly coupled contingency-table trio. Full-vector adjustment leaves a moderate negative estimate on MMLU-Pro, makes the TruthfulQA estimate much weaker, and leaves MuSiQue weak and uncertain. Double-fault should therefore be read as a shared-failure diagnostic within this constrained relationship.

On the 19 models whose recorded route IDs appear in all three rosters, using each benchmark's own unchanged primary item set, the qualitative findings persist: positive-gain rates are $10.0\%$, $4.3\%$, and $0.7\%$, strict--capability correlations are at least $0.96$, and best+mean double-fault remains negative, with the MuSiQue interval again including zero (Appendix Table~\ref{tab:shared-roster}). Aligning route membership is therefore insufficient to eliminate the observed contrasts; because items, prompts, answer formats, scoring, dates, and serving conditions still differ, we do not attribute them to question type or reasoning demands.

These results do not mean that error diversity is unimportant for aggregation. Instead, the audited observational statistics do not clearly separate useful error structure from member capability. Building on the classical accuracy--diversity trade-off and established contingency-table measures, our contribution is a metric-by-metric analysis of gain over the strongest constituent using explicit capability adjustment, three benchmarks, stronger-control sensitivity, common-route robustness, and model-level uncertainty. This evidence supports capability-aware interpretation of diversity-based signals.

\textbf{Practical recommendations.} The following workflow is a recommendation derived from the audit, not an experimentally validated selection algorithm.
\begin{enumerate}
\item Measure constituent capability first, including relevant cost and latency constraints where applicable \citep{chen2024frugalgpt}.
\item Do not select ensembles based solely on raw diversity-metric values.
\item Evaluate complementarity among candidate models of comparable capability using held-out development data.
\item Validate the intended selection and aggregation pipeline against a strongest individual model selected on development data and evaluated on untouched test data (Section~5.1).
\item Use pairwise co-failure as a diagnostic signal of shared failure, not as a guaranteed standalone ensemble selector.
\end{enumerate}

Unweighted majority voting is the classical rule whose diversity intuition we audit; it is not the only aggregation option. Alternatives include weighted voting, cost-aware routing \citep{chen2024frugalgpt,chen2025harnessing}, judge-based selection when a reference-free score is trusted \citep{kim2026judges}, published diversity-aware or competence-aware selection \citep{tekin-etal-2024-llm,cohen-etal-2026-dfpe}, and semantic aggregation for free-form answers. We do not evaluate or rank these alternatives here.

\section{Conclusion}

We audited five diversity-related measures as conditional correlates of majority-vote gain over the strongest constituent in heterogeneous LLM pools on MMLU-Pro ($30$ models, $453$ complete items), TruthfulQA ($22$ models, $400$ items; randomized option order), and MuSiQue ($20$ models, $578$ items). Three measurement findings follow. First, oracle complementarity is widespread, yet realized size-3 majority-vote improvement remains uncommon ($9.95\%$, $4.68\%$, and $0.61\%$). Second, several conventional statistics are capability-entangled; for strict diversity, the exact identity $\mathrm{Strict}=2(1-\overline{\mathrm{Acc}})-\mathrm{DF}$ makes much of the relationship structural, while the observed correlations ($\rho=+0.991/+0.988/+0.967$) remain pool-dependent. Third, best+mean controlled associations are metric- and benchmark-dependent. Double-fault is negative with model-subpool intervals excluding zero on MMLU-Pro ($-0.398$) and TruthfulQA ($-0.515$), whereas MuSiQue is weaker, uncertain, and roster-sensitive ($-0.175$); focal diversity provides a contrasting MuSiQue association ($-0.335$). Stronger capability control attenuates double-fault to $-0.284/-0.204/-0.065$. The contribution is an LLM-specific measurement audit that separates exact constraints from empirical association magnitudes.

\clearpage

\section{Limitations}

\begin{enumerate}
\item \textbf{Evaluation coverage and missingness.} Primary analyses use complete-case intersections of $453/500$ (MMLU-Pro), $400/500$ (TruthfulQA), and $578/600$ (MuSiQue) items and never score missing cells as incorrect. Coverage is incomplete and concentrated rather than missing completely at random: retained items tend to be easier than excluded ones on fixed-roster diagnostics. A focused MMLU-Pro available-item sensitivity uses, for each size-3 ensemble, only items observed for all three members among the frozen $500$. It yields controlled double-fault $-0.413$ versus $-0.398$ on the roster-wide $453$-item complete case, with per-ensemble denominators ranging from $460$ to $500$. These are different estimands; improved parsing coverage does not eliminate selection effects. On TruthfulQA, retained items are easier than dropped ones ($+0.193$ on 13 fully covered routes), and missing cells concentrate in a few routes (\texttt{gemini25\_flash\_lite}, \texttt{mistral\_small}, \texttt{glm47}). Exact positive-gain rates are therefore sensitive to the item set ($4.29\%$--$5.19\%$ across valid parse rules, denominators, and single-route removals in Appendix~\ref{app:historical-tqa}), and only their minority status is a robust conclusion.

\item \textbf{Unmatched model pools and setting dependence.} The three benchmarks use benchmark-specific rosters ($30$, $22$, and $20$ models) constructed under different route availability and inclusion criteria. Differences across benchmarks therefore confound task format with model composition, prompting, coverage, dates, serving conditions, answer format, and scoring. The three model counts are not independent sample sizes that can be compared directly. Results should not be attributed solely to question difficulty or answer format, nor extrapolated as roster-invariant laws of LLM ensembles. A 19-model exact route-ID intersection preserves the qualitative findings (Appendix Table~\ref{tab:shared-roster}) but does not match items, prompts, dates, answer format, scoring, or serving conditions.

\item \textbf{TruthfulQA protocol.} The randomized-option primary covers 22 available routes; seven routes from the original 29-route analysis pool were unavailable under their recorded identifiers and were not replaced. The final response set spans several dates (2026-11-21 to 2026-11-24 UTC). Provider routing was not pinned, and model versions or serving conditions may have changed. Responses truncated without a valid final answer were re-requested with larger output-token budgets ($1{,}024\rightarrow4{,}096\rightarrow8{,}192$), producing heterogeneous budgets across items and routes. Repeating sampled cells changed $3$--$4\%$ of parsed labels even at temperature $0$. The parser correction is gold-blind and uniform, but parse rules and denominators still move the exact positive-gain rate and the full-vector estimate (Section~5.5). The superseded fixed-correct-first evaluation cannot be repaired from its surviving cache because correct option position and parser error are confounded (Appendix~\ref{app:historical-tqa}).

\item \textbf{MuSiQue estimand and residual uncertainty.} MuSiQue analyses use correctness-count strict-majority voting as an oracle-style diagnostic over free-form correctness outcomes, not deployable free-text or semantic answer aggregation. The controlled double-fault point estimate is negative ($-0.175$), but its model-level interval includes zero. The conditional association is also roster-sensitive: on the 12-model pilot roster (a subset of the 20 primary routes, observed on all 600 items), the same best+mean estimator gives $+0.064$ on all 600 items and $+0.145$ on the 578 primary items, with model-level intervals that include zero ($[-0.241,+0.182]$ and $[-0.228,+0.257]$), and no size-3 ensemble beats its strongest member. We therefore do not generalize a statistically established residual co-failure association across all question formats.

\item \textbf{Dependent subsets and associational estimands.} Enumerated subsets reuse the same roster models and are highly overlapping; subset counts are not independent sample sizes. We report model-level unique-subpool intervals and leave-one-model-out ranges rather than subset-level significance tests. These intervals perturb model rosters but do not include item-sampling variability and should not be read as universal significance statements. Capability controls estimate conditional associations given observed member accuracy; they are not interventions that identify causal effects of diversity. Inclusion accuracy bands were pragmatic construction choices, not formally preregistered standards.

\item \textbf{Aggregation and prompt scope.} The study does not directly evaluate learned routers, weighted voting, judge-based aggregation, or semantic answer synthesis, and it does not rank those methods. Findings concern unweighted correctness-count majority voting under a shared zero-shot chain-of-thought prompting regime; other prompts or reasoning regimes may change both capability and error dependence. Practical interpretation of diversity signals is discussed in Section~6 without presenting a validated optimization algorithm. Concurrent work on quality-matched pools and accuracy-adjusted dependence \citep{chen2026combining,ali2026quantifying} is complementary; our scope remains diagnostic.
\end{enumerate}

\bibliography{references}

\appendix
\section{Model roster and answer parsing}

\subsection{Model roster and full-sample filtering statistics}

Parse rate and accuracy use the full 500-item samples. Full-sample accuracy is the number of correct predictions divided by 500; unparsed responses therefore contribute zero. MMLU-Pro inclusion requires parse rate \ensuremath{\geq}0.90 and full-sample accuracy in [0.40, 0.92]; the TruthfulQA randomized roster applies the parse-rate criterion only (Section~4).

\textbf{MMLU-Pro:} 30/30 routes pass the canonical filter.

\begin{table*}[!htbp]
\centering
\scriptsize
\caption*{Table A1. MMLU-Pro model roster and full-sample filtering statistics.}
\begin{tabularx}{\textwidth}{llrrr}
\toprule
Alias & OpenRouter model ID & Parse rate & Full-500 accuracy & Canonical roster \\
\midrule
\texttt{claude\_35\_haiku} & \texttt{anthropic/claude-3.5-haiku} & 99.8\% & 63.8\% & Yes \\
\texttt{claude\_37\_sonnet} & \texttt{anthropic/claude-3.7-sonnet} & 99.6\% & 82.4\% & Yes \\
\texttt{claude\_haiku} & \texttt{anthropic/claude-haiku-4.5} & 98.0\% & 79.0\% & Yes \\
\texttt{claude\_sonnet46} & \texttt{anthropic/claude-sonnet-4.6} & 95.2\% & 83.0\% & Yes \\
\texttt{ernie45} & \texttt{baidu/ernie-4.5-21b-a3b} & 97.0\% & 61.8\% & Yes \\
\texttt{seed16} & \texttt{bytedance-seed/seed-1.6} & 94.2\% & 78.8\% & Yes \\
\texttt{deepseek\_v3} & \texttt{deepseek/deepseek-chat-v3-0324} & 100.0\% & 79.2\% & Yes \\
\texttt{deepseek\_r1} & \texttt{deepseek/deepseek-r1} & 100.0\% & 84.6\% & Yes \\
\texttt{gemini25\_flash} & \texttt{google/gemini-2.5-flash} & 91.6\% & 69.8\% & Yes \\
\texttt{gemini25\_flash\_lite} & \texttt{google/gemini-2.5-flash-lite} & 94.8\% & 71.4\% & Yes \\
\texttt{gemini25\_pro} & \texttt{google/gemini-2.5-pro} & 99.8\% & 82.6\% & Yes \\
\texttt{gemma4\_31b} & \texttt{google/gemma-4-31b-it} & 93.6\% & 82.8\% & Yes \\
\texttt{llama31\_8b} & \texttt{meta-llama/llama-3.1-8b-instruct} & 100.0\% & 51.0\% & Yes \\
\texttt{llama33\_70b} & \texttt{meta-llama/llama-3.3-70b-instruct} & 100.0\% & 72.0\% & Yes \\
\texttt{llama4\_maverick} & \texttt{meta-llama/llama-4-maverick} & 94.2\% & 76.8\% & Yes \\
\texttt{llama4\_scout} & \texttt{meta-llama/llama-4-scout} & 95.0\% & 72.0\% & Yes \\
\texttt{minimax\_m25} & \texttt{minimax/minimax-m2.5} & 96.4\% & 72.2\% & Yes \\
\texttt{mistral\_large} & \texttt{mistralai/mistral-large} & 91.6\% & 74.0\% & Yes \\
\texttt{mistral\_small} & \texttt{mistralai/mistral-small-3.1-24b-instruct} & 97.2\% & 64.0\% & Yes \\
\texttt{kimi\_k2} & \texttt{moonshotai/kimi-k2-thinking} & 94.8\% & 80.2\% & Yes \\
\texttt{gpt41} & \texttt{openai/gpt-4.1} & 96.4\% & 79.4\% & Yes \\
\texttt{gpt41\_nano} & \texttt{openai/gpt-4.1-nano} & 92.6\% & 60.8\% & Yes \\
\texttt{gpt4o} & \texttt{openai/gpt-4o} & 99.0\% & 74.4\% & Yes \\
\texttt{gpt4o\_mini} & \texttt{openai/gpt-4o-mini} & 98.0\% & 49.4\% & Yes \\
\texttt{qwen35\_9b} & \texttt{qwen/qwen3.5-9b} & 94.4\% & 58.6\% & Yes \\
\texttt{qwen36\_plus} & \texttt{qwen/qwen3.6-plus} & 97.2\% & 85.4\% & Yes \\
\texttt{grok4\_fast} & \texttt{x-ai/grok-4-fast} & 100.0\% & 83.8\% & Yes \\
\texttt{grok41\_fast} & \texttt{x-ai/grok-4.1-fast} & 99.6\% & 85.0\% & Yes \\
\texttt{glm47} & \texttt{z-ai/glm-4.7} & 97.2\% & 70.6\% & Yes \\
\texttt{glm51} & \texttt{z-ai/glm-5.1} & 98.8\% & 75.0\% & Yes \\
\bottomrule
\end{tabularx}
\end{table*}

\textbf{TruthfulQA:} the primary roster contains 22 available routes (Table~A2); all pass parse rate \ensuremath{\geq}0.90 under randomized option order.

\textbf{MuSiQue:} The primary roster contains $20$ models with $578/600$ complete-case items (hop retention $299/175/104$ among complete-case items). Its fixed roster is listed in Table~A3.

\begin{table*}[!htbp]
\centering
\scriptsize
\caption*{Table A2. TruthfulQA primary roster under randomized option order (seed 2026) and the corrected parser.}
\begin{adjustbox}{max width=\textwidth}
\begin{tabular}{llrrr}
\toprule
Alias & OpenRouter model ID & Parse rate & Full-500 accuracy & Cells at 4,096/8,192 tokens \\
\midrule
\texttt{claude\_haiku45} & \texttt{anthropic/claude-haiku-4-5} & 100.0\% & 88.6\% & 0/0 \\
\texttt{claude\_sonnet46} & \texttt{anthropic/claude-sonnet-4.6} & 100.0\% & 89.0\% & 0/0 \\
\texttt{deepseek\_v3} & \texttt{deepseek/deepseek-chat-v3-0324} & 100.0\% & 84.2\% & 0/0 \\
\texttt{gemini25\_flash} & \texttt{google/gemini-2.5-flash} & 99.8\% & 81.8\% & 63/2 \\
\texttt{gemini25\_flash\_lite} & \texttt{google/gemini-2.5-flash-lite} & 92.8\% & 79.4\% & 93/4 \\
\texttt{gemini25\_pro} & \texttt{google/gemini-2.5-pro} & 100.0\% & 92.6\% & 480/13 \\
\texttt{gemma4\_31b} & \texttt{google/gemma-4-31b-it} & 100.0\% & 89.0\% & 0/0 \\
\texttt{llama31\_8b} & \texttt{meta-llama/llama-3.1-8b-instruct} & 98.6\% & 57.6\% & 3/0 \\
\texttt{llama33\_70b} & \texttt{meta-llama/llama-3.3-70b-instruct} & 100.0\% & 79.0\% & 0/0 \\
\texttt{llama4\_maverick} & \texttt{meta-llama/llama-4-maverick} & 100.0\% & 79.6\% & 0/0 \\
\texttt{llama4\_scout} & \texttt{meta-llama/llama-4-scout} & 100.0\% & 75.2\% & 0/0 \\
\texttt{minimax\_m25} & \texttt{minimax/minimax-m2.5} & 100.0\% & 82.8\% & 12/1 \\
\texttt{mistral\_large} & \texttt{mistralai/mistral-large} & 98.0\% & 87.6\% & 213/44 \\
\texttt{mistral\_small} & \texttt{mistralai/mistral-small-3.1-24b-instruct} & 93.4\% & 72.0\% & 0/0 \\
\texttt{kimi\_k2} & \texttt{moonshotai/kimi-k2-thinking} & 100.0\% & 91.0\% & 0/0 \\
\texttt{gpt41} & \texttt{openai/gpt-4.1} & 100.0\% & 87.2\% & 0/0 \\
\texttt{gpt41\_nano} & \texttt{openai/gpt-4.1-nano} & 99.4\% & 69.6\% & 0/0 \\
\texttt{gpt4o} & \texttt{openai/gpt-4o} & 100.0\% & 84.6\% & 0/0 \\
\texttt{gpt4o\_mini} & \texttt{openai/gpt-4o-mini} & 100.0\% & 69.6\% & 0/0 \\
\texttt{qwen35\_9b} & \texttt{qwen/qwen3.5-9b} & 98.6\% & 82.2\% & 179/14 \\
\texttt{glm47} & \texttt{z-ai/glm-4.7} & 96.4\% & 90.6\% & 94/7 \\
\texttt{glm51} & \texttt{z-ai/glm-5.1} & 99.2\% & 90.8\% & 90/10 \\
\bottomrule
\end{tabular}
\end{adjustbox}
\par\smallskip
{\scriptsize \textbf{Roster note.} All 22 routes meet the randomized inclusion rule (parse rate \ensuremath{\geq}0.90). No additional post hoc accuracy exclusion is applied; \texttt{gemini25\_pro} (92.6\%) is the only route above the original 92\% construction bound, with a leave-route-out sensitivity in Appendix~\ref{app:historical-tqa}. Seven original analysis routes were unavailable under their recorded identifiers and were not replaced: \texttt{claude\_35\_haiku}, \texttt{claude\_37\_sonnet}, \texttt{deepseek\_r1}, \texttt{ernie45}, \texttt{seed16}, \texttt{grok4\_fast}, and \texttt{grok41\_fast}. Route identifiers are reported exactly as served; in particular, the Claude Haiku IDs differ across MMLU-Pro and TruthfulQA.\par}
\end{table*}

\begin{table*}[!htbp]
\centering
\scriptsize
\caption*{Table A3. MuSiQue primary model roster.}
\begin{adjustbox}{max width=\textwidth}
\begin{tabular}{llll}
\toprule
Alias & OpenRouter model ID & Alias & OpenRouter model ID \\
\midrule
\texttt{gemma4\_31b} & \texttt{google/gemma-4-31b-it} & \texttt{claude\_sonnet46} & \texttt{anthropic/claude-sonnet-4.6} \\
\texttt{kimi\_k2} & \texttt{moonshotai/kimi-k2-thinking} & \texttt{gpt41} & \texttt{openai/gpt-4.1} \\
\texttt{llama4\_maverick} & \texttt{meta-llama/llama-4-maverick} & \texttt{deepseek\_v3} & \texttt{deepseek/deepseek-chat-v3-0324} \\
\texttt{gemini25\_flash} & \texttt{google/gemini-2.5-flash} & \texttt{llama4\_scout} & \texttt{meta-llama/llama-4-scout} \\
\texttt{gpt4o} & \texttt{openai/gpt-4o} & \texttt{llama33\_70b} & \texttt{meta-llama/llama-3.3-70b-instruct} \\
\texttt{mistral\_small} & \texttt{mistralai/mistral-small-3.1-24b-instruct} & \texttt{gpt4o\_mini} & \texttt{openai/gpt-4o-mini} \\
\texttt{llama31\_8b} & \texttt{meta-llama/llama-3.1-8b-instruct} & \texttt{qwen35\_9b} & \texttt{qwen/qwen3.5-9b} \\
\texttt{gpt41\_nano} & \texttt{openai/gpt-4.1-nano} & \texttt{gemini25\_flash\_lite} & \texttt{google/gemini-2.5-flash-lite} \\
\texttt{glm51} & \texttt{z-ai/glm-5.1} & \texttt{mistral\_large} & \texttt{mistralai/mistral-large} \\
\texttt{gemini25\_pro} & \texttt{google/gemini-2.5-pro} & \texttt{qwen36\_plus} & \texttt{qwen/qwen3.6-plus} \\
\bottomrule
\end{tabular}
\end{adjustbox}
\par\smallskip
{\scriptsize \textbf{Roster note.} The fixed 20-route roster comprises the 12-route pilot panel plus eight additional original routes selected using historical MMLU-Pro capability and model-family coverage. The expansion roster was fixed before inference on those eight routes, and no MuSiQue-specific accuracy band was applied post hoc.\par}
\end{table*}

\subsection{Prompting, inference, and parsing protocol}

\textbf{Benchmarks and sampling.} MMLU-Pro is loaded from \texttt{TIGER-Lab/MMLU-Pro} (test split; 12,032 items). We sample 500 row indices uniformly without replacement using Python's \texttt{random.sample} after setting seed 42; no category stratification is used. TruthfulQA is loaded from \texttt{truthfulqa/truthful\_qa} (\texttt{multiple\_choice} configuration, validation split), from which 500 items are sampled by the same procedure and seed. We use the \texttt{mc1\_targets} choices and the unique choice labeled 1 as the TruthfulQA answer. The sampled-question files store the original dataset row index, local question ID, question text, options, and answer label; all 500 TruthfulQA mappings were verified against the Hugging Face validation-row order.

\textbf{Initial prompt templates.} Both benchmarks use zero-shot chain-of-thought multiple-choice prompts with no demonstrations. \textbf{No system prompt was used:} every request contained one user message only. In the templates below, \texttt{{question}} is replaced by the benchmark question and \texttt{{lettered\_options}} by one option per line in \texttt{A. option text}, \texttt{B. option text}, ... format. Individual benchmark questions are not part of the template.

MMLU-Pro:

\begin{quote}
\begin{minipage}{0.96\linewidth}
\ttfamily\footnotesize\raggedright
Answer the following multiple choice question.\par
Think step by step, then provide your final answer.\par
At the very end, write your answer in this exact format:\par
ANSWER: X\par
(where X is the letter of the correct option)\par
\smallskip
Question: \{question\}\par
\smallskip
Options:\par
\{lettered\_options\}\par
\smallskip
Let's think step by step:\par
\end{minipage}
\end{quote}

TruthfulQA:

\begin{quote}
\begin{minipage}{0.96\linewidth}
\ttfamily\footnotesize\raggedright
Answer the following question truthfully.\par
Think step by step, then provide your final answer.\par
At the very end, write your answer in this exact format:\par
ANSWER: X\par
(where X is the letter of the correct option)\par
\smallskip
Question: \{question\}\par
\smallskip
Options:\par
\{lettered\_options\}\par
\smallskip
Let's think step by step:\par
\end{minipage}
\end{quote}

Table~A4 summarizes the MMLU-Pro inference and missing-response recovery settings. TruthfulQA's randomized-option protocol follows separately because its failure-based token ladder differs.

\begin{table}[!htbp]
\centering
\scriptsize
\caption*{Table A4. MMLU-Pro inference and missing-response recovery settings.}
\begin{tabularx}{\columnwidth}{>{\raggedright\arraybackslash}p{0.18\columnwidth}>{\raggedright\arraybackslash}X>{\raggedright\arraybackslash}X}
\toprule
Setting & Initial evaluation & Unparsed-response retry \\
\midrule
message roles & one user message; no system message & one user message; no system message \\
decoding & temperature 0.0; \texttt{top\_p}, stop sequences, penalties, and request seed unset (provider defaults) & same \\
max output tokens & 1,024, except 8,192 for five designated reasoning routes listed below & 2,048, except 4,096 for \texttt{qwen35\_9b} \\
request timeout & 90 s for parser group 1; SDK/provider default otherwise & SDK/provider default \\
API-error attempts & up to 3 per prompt; 3/6-s backoff for parser group 1 and 4/8-s otherwise & up to 3 per prompt; 4/8-s backoff \\
parser source & response \texttt{content} & \texttt{content}, then provider \texttt{reasoning} field as fallback \\
\bottomrule
\end{tabularx}
\end{table}

The five MMLU-Pro routes assigned 8,192 initial output tokens were \texttt{deepseek\_r1}, \texttt{kimi\_k2}, \texttt{gemini25\_pro}, \texttt{gemini25\_flash}, and \texttt{gemini25\_flash\_lite}.

\textbf{Multiple-choice parsing.}

Matching ignores case. Labels outside the available option range are rejected. Patterns run in order, using the last match from each.

\textbf{MMLU-Pro parser group 1 (eight routes).} In order: \texttt{ANSWER: X}, \texttt{answer: X}, \texttt{The answer is X}, \texttt{Final answer: X}, and a response consisting only of one label.

\textbf{MMLU-Pro parser group 2 (remaining routes).} First remove complete \texttt{<think>...</think>} blocks; then check \texttt{ANSWER: X}, \texttt{answer: X}, boxed labels, bold \texttt{Answer: X}, bold labels, \texttt{The [correct] answer is X}, the Chinese prefixes encoded by U+9009/U+62E9 (“selection”), U+7B54/U+6848 (“answer”), or U+6B63/U+786E/U+7B54/U+6848 (“correct answer”), a line beginning with \texttt{X.} or \texttt{X)}, and a response consisting only of one label.

\textbf{TruthfulQA.} First remove complete \texttt{<think>...</think>} blocks; then check \texttt{ANSWER: X}, \texttt{answer: X}, boxed labels, bold labels, \texttt{The [correct] answer is X}, and a response consisting only of one label. If no valid label is found, the record is stored with no prediction and \texttt{parse\_success=False}; it is not scored as a parsed prediction.

\textbf{TruthfulQA randomized-option primary.} The 500 frozen items are displayed with seed-2026 option permutations. The gold position is balanced within each choice-count group, and every prompt is hash-checked against the frozen record before sending. The prompt is the TruthfulQA template above, sent as one user message at temperature 0. The MMLU-Pro retry prompts below are not used. The output budget follows a failure-based ladder that is identical for all routes. Requests start at 1,024 tokens. A response that is truncated without a parseable label moves the item to 4,096 and then 8,192 tokens. Transport and provider errors repeat the identical request. Other unparsed HTTP-200 responses receive one same-budget retry. For attempts after the first, the parser reads the provider \texttt{reasoning} field when \texttt{content} is empty. The selected answer is the first parsed attempt. The final response set was collected from 2026-11-21 to 2026-11-24 UTC under unpinned providers.

\textbf{Corrected TruthfulQA parser.} The ordered patterns above take the first character after a cue, so a cue followed by a word (e.g., \texttt{Final answer: ANSWER: C} or \texttt{The answer is Gold}) can be misread as A or another letter. We apply a uniform, gold-blind correction. (i) A parse whose label is a standalone token is kept. (ii) If the label was captured from inside a word, the same patterns are re-run requiring a standalone label, but only when the response finished normally. (iii) Truncated responses with such a capture, and completed responses with no standalone label, are left missing. (iv) No parse is created for a response that was unparsed, so the selected attempt never changes. Of 10,891 parsed cells, 284 are re-parsed, 233 change label, and 10 become missing. The final complete case contains 400 items.

\textbf{MMLU-Pro unparsed-response prompts.} MMLU-Pro responses with \texttt{parse\_success=False} use the following three templates in order. Options in Prompts 1–2 use \texttt{A. option text}; Prompt 3 uses \texttt{(A) option text}.

Prompt 1:

\begin{quote}
\begin{minipage}{0.96\linewidth}
\ttfamily\footnotesize\raggedright
Answer the following multiple choice question.\par
Think step by step, then provide your final answer.\par
At the very end, write your answer in this exact format:\par
ANSWER: X\par
(where X is the letter of the correct option)\par
\smallskip
Question: \{question\}\par
\smallskip
Options:\par
\{lettered\_options\}\par
\smallskip
Let's think step by step:\par
\end{minipage}
\end{quote}

Prompt 2:

\begin{quote}
\begin{minipage}{0.96\linewidth}
\ttfamily\footnotesize\raggedright
Question: \{question\}\par
\smallskip
\{lettered\_options\}\par
\smallskip
Which option is correct? Reply with just the letter.\par
Your answer (one letter only):\par
\end{minipage}
\end{quote}

Prompt 3:

\begin{quote}
\begin{minipage}{0.96\linewidth}
\ttfamily\footnotesize\raggedright
Select the correct answer.\par
\smallskip
\{question\}\par
\smallskip
\{parenthesized\_options\}\par
\smallskip
Answer with only one letter like: A\par
\end{minipage}
\end{quote}

For each retry prompt, we first parse response \texttt{content} using this ordered cascade: \texttt{ANSWER: X}, \texttt{answer: X}, \texttt{Answer: X}, bold \texttt{X}, \texttt{The answer is X}, \texttt{correct answer is X}, a standalone \texttt{X}, and a final valid option letter. If content parsing fails, the same parser is applied to the provider \texttt{reasoning} field. A successful parse stops the sequence immediately; otherwise the next prompt is tried. After all three prompts fail, the entry remains unparsed.

\textbf{Unparsed-response scope.} The additional prompts apply only to responses with \texttt{parse\_success=False} from routes below the parse-rate threshold. Parsed predictions are not regenerated or replaced. After a successful parse, only the targeted missing record is updated. The \texttt{retry\_parsing.py} script implements target selection, the three-prompt schedule, parser fallback, and merge checks.

\textbf{MuSiQue protocol.} MuSiQue uses \texttt{MuSiQue-Ans v1.0} development items. The frozen sample contains $600$ hop-stratified items. Every prompt includes all paragraphs in the public item record, including supporting and distractor paragraphs, with titles and text, but no gold answer, alias, or supporting-paragraph label. Each request uses one user message, temperature $0$, \texttt{top\_p} unset, and 2,048 maximum output tokens. The exact instruction is:
\begin{quote}
\begin{minipage}{0.96\linewidth}
\ttfamily\footnotesize\raggedright
Answer the multi-hop question using ONLY the paragraphs below.\par
Do not invent facts not supported by the paragraphs.\par
Think step by step, then put the concise final answer on the last line in this exact format:\par
Final answer: <answer>\par
\smallskip
Question: \{question\}\par
\smallskip
Paragraphs:\par
\{paragraphs\}
\end{minipage}
\end{quote}
The parser removes complete \texttt{<think>} blocks and requires an explicit final-answer marker. Binary correctness is maximum normalized exact match over the official answer and aliases; token F1 is stored but not thresholded. Unparsed cells have null correctness and are excluded by the complete-case rule. The fixed 20-model roster is listed in Table~A3; the multiple-choice accuracy-band filter is not applied to MuSiQue.

\textbf{Voting rules.} The canonical correct-count rule marks an item correct only when more than half the subset members are correct; exact even-size splits are incorrect. The answer-level plurality robustness check requires the correct option to be the unique most-voted answer; any tie for the highest answer count is marked incorrect. For single-label multiple choice, these rules are identical for subset sizes 2–3 and can differ at size 4. For free text, distinct correct strings can split; MuSiQue's correctness-count result remains a diagnostic rather than answer-level voting.

\FloatBarrier
\section{Model-level resampling procedure}

Enumerated subsets heavily overlap because they reuse the same roster models within each benchmark ($M=30$ on MMLU-Pro, $22$ on TruthfulQA, $20$ on MuSiQue). We therefore avoid subset-level $p$-values and use model-level unique-subpool resampling for robustness intervals on partial Spearman associations ($B=1{,}000$; NumPy Generator seed~$0$). Pseudocode:

\begin{enumerate}
\item Draw a multiset of model indices of size $M$ uniformly with replacement.
\item Deduplicate to a set $U$ of unique drawn models. If $|U|<8$, discard the replicate.
\item Retain only subsets whose members are all in $U$ (subsets that use any non-drawn model are dropped). If fewer than 50 subsets remain, discard the replicate.
\item On the retained subsets, recompute the partial Spearman of the audited measure (double-fault unless stated otherwise) with gain given {best, mean} (rank-space residualization, same as the main tables).
\item The 95\% interval is the 2.5th–97.5th percentile of valid replicate statistics.
\end{enumerate}

Because step 2 deduplicates draws, duplicate multiplicities do not reweight the statistic. The procedure changes the model roster through random subpools rather than using a conventional cluster bootstrap. We therefore report model-level resampling robustness intervals, not bootstrap confidence intervals. Leave-one-model-out estimates drop all subsets containing a held-out model and recompute the same partial association.

\subsection{Residual identity verification}

Numerically, using the raw (unranked) measures, $\max|\mathrm{double\text{-}fault}_{\mathrm{res}}+\tfrac12\mathrm{disagree}_{\mathrm{res}}|$ is at machine precision ($\sim 3\times 10^{-16}$) after linearly removing best and mean member accuracy at size 3. This matches the algebraic prediction of Section~3.1: the residuals are perfectly collinear (Pearson $r=-1.000$) with slope $-1/2$. Appendix Table~\ref{tab:five-metric-associations} instead uses rank-space residualization and reports the association with majority-vote gain analyzed in Section~5.4, rather than the raw-space collinearity.

\FloatBarrier
\section{Canonical associations and robustness}
\label{app:five-metric}

\begin{table}[H]
\centering
\begin{adjustbox}{max width=\linewidth}
\small
\begin{tabular}{lrrrrrr}
\toprule
& \multicolumn{3}{c}{Raw Spearman} & \multicolumn{3}{c}{Best+mean partial} \\
\cmidrule(lr){2-4}\cmidrule(lr){5-7}
Metric & MMLU & TQA & MuSiQue & MMLU & TQA & MuSiQue \\
\midrule
Strict diversity & $-0.675$ & $-0.571$ & $-0.425$ & $0.324$ & $0.239$ & $0.006$ \\
Pairwise disagreement & $-0.690$ & $-0.621$ & $-0.585$ & $0.290$ & $0.246$ & $-0.130$ \\
Double-fault & $-0.482$ & $-0.299$ & $-0.182$ & $-0.398$ & $-0.515$ & $-0.175$ \\
Jaccard error similarity & $0.517$ & $0.544$ & $0.518$ & $-0.209$ & $-0.179$ & $0.163$ \\
Focal diversity & $-0.486$ & $-0.499$ & $-0.598$ & $0.106$ & $0.085$ & $-0.335$ \\
\bottomrule
\end{tabular}
\end{adjustbox}
\caption{Size-3 associations between each diversity measure and correctness-count gain. The right panel residualizes ranks on the ensemble's best and mean member accuracy. These are conditional observational associations. Strict diversity equals disagreement plus double-fault by construction.}
\label{tab:five-metric-associations}
\end{table}

\begin{table}[H]
\centering
\small
\begin{adjustbox}{max width=\linewidth}
\begin{tabular}{lrrl}
\toprule
Benchmark & Best+mean & Full vector & Full-vector interval \\
\midrule
MMLU-Pro & $-0.398$ & $-0.284$ & $[-0.461,-0.049]$ \\
TruthfulQA & $-0.515$ & $-0.204$ & $[-0.372,-0.053]$ \\
MuSiQue & $-0.175$ & $-0.065$ & $[-0.310,+0.242]$ \\
\bottomrule
\end{tabular}
\end{adjustbox}
\caption{Control sensitivity of the size-3 double-fault association. Full vector controls for the ranked sorted accuracies of all three members. Intervals are unique-model subpool percentiles (seed 0), not confidence intervals over items.}
\label{tab:full-vector-control}
\end{table}

\textbf{Focal-diversity interpretation at size~3.} Let $f_i$ be focal member $i$'s failure rate, $d_{ij}$ and $d_{ik}$ its two pairwise double-fault rates, and $F$ the all-three-failure rate. The implemented definition reduces exactly to $\rho_i=1-(d_{ij}+d_{ik}+F)/(f_i+d_{ij}+d_{ik})$. Focal diversity therefore contains the all-three-failure term in addition to pairwise information. Because $F$ also enters Equation~\ref{eq:size3-gain}, the controlled MuSiQue estimate ($-0.335$) is a conditional association among the audited metrics, not evidence of an independent higher-order mechanism or causal effect.

\textbf{Held-out best-member selection (size~3).} Using the protocol in Section~5.1 (seeds $0$--$19$; select the strongest member on a random half of complete-case items and evaluate on the complementary half), positive-gain rates are higher than the in-sample rates on every benchmark (Table~\ref{tab:heldout-size3}). This does not mean voting performs better out of sample: the development-selected member need not be strongest on the evaluation half, so the realized comparison baseline is weaker than the in-sample maximum.

\subsection{TruthfulQA protocol robustness and correction}
\label{app:historical-tqa}

\textbf{Robustness of the randomized primary.} Table~\ref{tab:tqa-robustness} recomputes the size-3 results under alternative parse rules, denominators, and rosters. The best+mean double-fault estimate stays between $-0.49$ and $-0.54$ in every row, and every reported model-level interval excludes zero. The positive-gain rate stays a small minority (4.29\%--5.19\% for valid parse rules). The full-vector estimate is the least stable quantity: its interval includes zero under the independent parser and when every truncated response is treated as missing, and it is near zero under per-ensemble denominators. We therefore do not claim that the full-vector estimate is stable.

\begin{table*}[t]
\centering
\begin{adjustbox}{max width=0.8\textwidth}
\small
\begin{tabular}{lrrll}
\toprule
Panel & Items & Gain$>0$ & Best+mean DF & Full-vector DF \\
\midrule
Primary (corrected parser) & 400 & 4.68\% & $-0.515$ $[-0.656,-0.241]$ & $-0.204$ $[-0.372,-0.053]$ \\
Independent parser & 365 & 4.87\% & $-0.520$ $[-0.653,-0.276]$ & $-0.157$ $[-0.337,+0.018]$ \\
Truncated responses missing & 361 & 4.55\% & $-0.489$ $[-0.633,-0.218]$ & $-0.123$ $[-0.310,+0.058]$ \\
Without \texttt{gemini25\_pro} (21) & 400 & 5.19\% & $-0.501$ $[-0.661,-0.174]$ & $-0.226$ $[-0.394,-0.051]$ \\
Without \texttt{mistral\_small} (21) & 426 & 4.44\% & $-0.509$ $[-0.674,-0.212]$ & $-0.212$ $[-0.402,-0.001]$ \\
Per-ensemble available items & 421--500 & 4.29\% & $-0.526$ & $-0.038$ \\
\bottomrule
\end{tabular}
\end{adjustbox}
\caption{TruthfulQA size-3 sensitivity (22 routes unless noted; 1,540 or 1,330 ensembles). Intervals are unique-model subpool percentiles (seed 0). The per-ensemble row uses the items observed for each triplet. Rows span parser, truncation, roster, and denominator choices; the full-vector estimate is less stable than the best+mean estimate.}
\label{tab:tqa-robustness}
\end{table*}

\textbf{Repeatability, budget, and inclusion.} Re-requesting 566 sampled cells at temperature 0 changed $3.1\%$ of parsed labels between the cached response and a 1,024-token rerun (553 cells). It changed $3.4\%$ between 1,024- and 4,096-token reruns, and $3.3\%$ on 424 cells where the budget was not binding. On completed cells, observed drift is therefore similar across these comparisons; this does not identify the budget effect for responses that truncate at 1,024 tokens. Budget escalation was applied identically across routes but was concentrated in reasoning-heavy routes (Table~A2). Removing \texttt{gemini25\_pro}, the route above the original construction bound, does not change the conclusions (Table~\ref{tab:tqa-robustness}).

\textbf{Superseded fixed-correct-first evaluation.} A prior public manuscript used a 29-route, 458-item TruthfulQA panel in which every gold option appeared in position A. That design confounds answer position with correctness. It was also affected by a cue-plus-word parser defect that could map text inside a word to option A. Because raw responses survive for only 1.1\% of final cells, the panel cannot be repaired by re-parsing and is not used as evidence in this paper. This correction notice is retained because the valid randomized evaluation replaces that earlier public evidence.

\subsection{MuSiQue coverage and voting estimand}

MuSiQue's frozen sample contains $600$ hop-stratified questions ($300/180/120$ two-/three-/four-hop). The primary complete-case panel retains $578$ items ($299/175/104$). Primary analyses use correctness-count strict-majority voting: an item is majority-correct iff more than half the members are correct. This is an oracle-style diagnostic over free-form correctness outcomes, not deployable free-text answer voting. An answer-level plurality analysis using normalized strings is positive for only $3/1140$ ensembles ($0.26\%$) and is not the primary estimand. Controlled double-fault on the primary panel is $-0.175$ with model-level interval $[-0.335,+0.050]$ (includes zero); leave-one-model-out remains negative ($[-0.212,-0.103]$). Hop strata describe coverage and are not independent replications. We did not validate or report hop-level diversity--gain associations. The 20-model aliases are listed in Appendix~A.1.

\subsection{Exact shared-route robustness check}
\label{app:shared-roster}

Table~\ref{tab:shared-roster} repeats the size-3 analysis on the 19 exact OpenRouter route IDs present in all three primary rosters (Tables~A1--A3). Each benchmark retains its own primary items, so only the roster is aligned. All 969 triplets per benchmark remain oracle-positive. Shared-roster double-fault intervals are $[-0.649,-0.137]$, $[-0.673,-0.162]$, and $[-0.362,+0.091]$ for MMLU-Pro, TruthfulQA, and MuSiQue. Matching routes does not match items, prompts, answer formats, scoring, or serving conditions.

\begin{table}[!htbp]
\centering
\small
\begin{tabular}{@{}lrrr@{}}
\toprule
Benchmark & Gain$>0$ P/S & Strict P/S & DF P/S \\
\midrule
MMLU-Pro & 9.95/10.01\% & .991/.989 & $-.398/-.479$ \\
TruthfulQA & 4.68/4.33\% & .988/.989 & $-.515/-.521$ \\
MuSiQue & 0.61/0.72\% & .967/.962 & $-.175/-.136$ \\
\bottomrule
\end{tabular}
\caption{Size-3 primary (P) and shared-19 (S) results. Strict correlates strict diversity with one minus mean accuracy; DF is the best+mean partial Spearman association. Each shared-roster row uses 969 overlapping triplets.}
\label{tab:shared-roster}
\end{table}

\subsection{Exploratory held-out predictive-validity audit}
\label{app:predictive-validity}

We ask whether a diversity metric improves out-of-item prediction of size-3 gain beyond constituent capability. For each benchmark and deterministic seed $0$--$19$, complete-case items are randomly divided into development set $A$ (the first $\lfloor n/2\rfloor$ permuted items) and held-out set $B$. All capability and diversity features are computed on $A$. Let $j_A^\star$ be the first member in canonical subset order attaining the largest accuracy on $A$. The training and evaluation targets are
\begin{align*}
G_A&=\mathrm{Acc}_{\mathrm{MV},A}-\mathrm{Acc}_{A}(j_A^\star),\\
G_B&=\mathrm{Acc}_{\mathrm{MV},B}-\mathrm{Acc}_{B}(j_A^\star).
\end{align*}
Thus, $B$ is never used to select or reselect the comparison member. The $A$-selected member is strictly less accurate than the $B$-optimal member in $16.38\%$ of MMLU-Pro's $81{,}200$, $14.19\%$ of TruthfulQA's $30{,}800$, and $12.65\%$ of MuSiQue's $22{,}800$ triplet--split pairs. Accordingly, $G_B$ is gain over a development-selected member, not the oracle-best member on $B$; the exact size-3 identity in Equation~\ref{eq:size3-gain} does not apply unchanged to $G_B$ when those members differ.

For each metric separately, we compare a capability-only predictor with the same predictor plus the metric. The two capability specifications are best and mean member accuracy on $A$ (B+M) and the full sorted vector of the three member accuracies on $A$ (Full). Features are standardized using $A$-derived rows only, and both models use the fixed specification \texttt{Ridge(alpha=1.0, fit\_intercept=True, solver=svd)} with $G_A$ as the training target. Predictions for the same triplets are evaluated against $G_B$, and $\Delta R^2=R^2_{\mathrm{capability+metric}}-R^2_{\mathrm{capability}}$ is paired within split, target, and subset.

\begin{table*}[!htbp]
\centering
\small
\begin{adjustbox}{max width=\textwidth}
\begin{tabular}{lrrrrrr}
\toprule
& \multicolumn{2}{c}{MMLU-Pro} & \multicolumn{2}{c}{TruthfulQA} & \multicolumn{2}{c}{MuSiQue} \\
\cmidrule(lr){2-3}\cmidrule(lr){4-5}\cmidrule(lr){6-7}
Metric & B+M & Full & B+M & Full & B+M & Full \\
\midrule
Strict diversity & $+0.0262$ & $-0.0027$ & $+0.0605$ & $-0.0072$ & $+0.0098$ & $-0.0022$ \\
Pairwise disagreement & $+0.0239$ & $-0.0037$ & $+0.0551$ & $-0.0101$ & $+0.0090$ & $-0.0024$ \\
Double-fault & $+0.0233$ & $-0.0038$ & $+0.0536$ & $-0.0107$ & $+0.0087$ & $-0.0024$ \\
Jaccard error similarity & $+0.0059$ & $-0.0035$ & $+0.0005$ & $-0.0084$ & $+0.0046$ & $-0.0024$ \\
Focal diversity & $+0.0055$ & $+0.0009$ & $+0.0033$ & $-0.0019$ & $-0.0002$ & $+0.0005$ \\
\midrule
Capability-only $R^2$ & $0.5382$ & $0.5952$ & $0.3939$ & $0.5096$ & $0.5916$ & $0.6498$ \\
\bottomrule
\end{tabular}
\end{adjustbox}
\caption{Held-out predictive validity at size~3. Metric rows report mean $\Delta R^2$ across 20 deterministic item splits relative to the corresponding capability-only baseline; the final row reports its mean held-out $R^2$. B+M uses best and mean member accuracy, and Full uses the sorted three-member accuracy vector. Strict diversity, disagreement, and double-fault span one linear incremental signal once mean accuracy is included; their small Ridge differences are representation-sensitive rather than independent predictive discoveries. Splits reuse items and triplets overlap in models, so the 20 values are not independent replications.}
\label{tab:predictive-validity}
\end{table*}

The positive B+M increments for strict diversity, disagreement, and double-fault therefore describe one algebraically shared direction: unpenalized OLS predictions from the three parameterizations agree within $2.64\times10^{-15}$, whereas Ridge changes slightly with feature scaling and parameterization. The main improvement does not remain material under Full. Seed-level signs are mixed; for example, the Full MuSiQue focal increment is positive in $16/20$ splits but averages only $+0.0005$ and ranges from $-0.0077$ to $+0.0054$.

This analysis tests transfer across held-out items for the same overlapping triplets in fixed model rosters. The reused item splits and shared constituent models preclude independent-replication or subset-level significance interpretations. It does not evaluate unseen-model generalization, prospective ensemble selection, routing, or a causal mechanism, and the structural coupling among capability, double-fault, and gain remains part of the predictive signal.

\end{document}